\documentclass[11pt,a4paper]{article}
\usepackage[hyperref]{acl2019}
\usepackage{times}
\usepackage{latexsym}
\usepackage{graphicx}
\usepackage{multirow}
\usepackage{url}
\usepackage{amsmath}
\usepackage{todonotes}
\usepackage{booktabs}
\usepackage{comment}

\usepackage{url}

\aclfinalcopy % Uncomment this line for the final submission
\title{Embedding time expressions for deep temporal ordering models}
\author{Tanya Goyal \and Greg Durrett \\
  Department of Computer Science \\
  The University of Texas at Austin \\
  {\tt tanyagoyal@utexas.edu, gdurrett@cs.utexas.edu}}

\date{}

\begin{document}
\maketitle
\begin{abstract}
Data-driven models have demonstrated state-of-the-art performance in inferring the temporal ordering of events in text. However, these models often overlook explicit temporal signals, such as dates and time windows. Rule-based methods can be used to identify the temporal links between these time expressions (timexes), but they fail to capture timexes' interactions with events and are hard to integrate with the distributed representations of neural net models. In this paper, we introduce a framework to infuse temporal awareness into such models by learning a pre-trained model to embed timexes. We generate synthetic data consisting of pairs of timexes, then train a character LSTM to learn embeddings and classify the timexes' temporal relation. We evaluate the utility of these embeddings in the context of a strong neural model for event temporal ordering, and show a small increase in performance on the MATRES dataset and more substantial gains on an automatically collected dataset with more frequent event-timex interactions.\footnote{Data and code are available at \url{https://github.com/tagoyal/Temporal-event-ordering}}
\end{abstract}

\section{Introduction}
Understanding the temporal ordering of events in a document is an important component of document understanding and plays an integral role in tasks such as timeline creation \cite{do2012joint}, temporal question answering \cite{llorens2015semeval} and causality inference \cite{mostafazadeh2016caters, ning2018joint}. %However, inferring the event temporal order is challenging as it often disagrees with the narrative order in text, and instead moves back and forth along the time dimension. Therefore, a key component of temporal understanding are time expressions that help anchor events to the time axis by providing explicit time signals in text i.e. date mentions and time windows.
Inferring temporal event order is challenging as it often disagrees with the narrative order in text. Past work on temporal relation extraction has exploited cues such as global constraints on the temporal graph structure \cite{Bramsen,Chambers:2008, ning2017structured}, world knowledge \cite{ning2018improving}, grouping of events \cite{tourille2017neural}, or fusing these cues more effectively with deep models \cite{meng2017temporal, cheng2017classifying}. One key component of temporal understanding is time expressions (timexes) that help anchor events to the time axis, but few recent systems effectively use the knowledge derivable from time expressions in their models. They either give timexes no special treatment \cite{ning2017structured} or rely on rule-based post-processing modules to remove inconsistencies with explicit timexes \cite{chambers2014dense,meng2017temporal}. 

In this work, we address this shortcoming by introducing a framework for including rich representations of timexes in neural models. These models implicitly capture some information via word embeddings \cite{mikolov2013distributed,pennington2014glove} or contextualized embeddings such as ELMo \cite{Peters:2018}. However, these embeddings do not encode the full richness of temporal information needed for this task. For example, these systems fail to infer the correct event relation in the following sentence: \textit{He visited France in 1992 and went to Germany in 1963.}~partially because the dates \emph{1992} and \emph{1963} do not have temporally-informed embeddings.

We devise a method for embedding timexes that more explicitly reflects their temporal status.
Specifically, we sample pairs of time expressions from synthetic data, train character LSTM models to encode these time expressions and classify their temporal ordering. Due to the amount and type of data they are trained on, these time embeddings will naturally capture the temporal ordering of events in standard text and generalize to things like unseen timex values.

We incorporate these embeddings into neural models for temporal relation extraction. When used in an improved version of the model from \citet{cheng2017classifying}, we show a small improvement in performance on the benchmark MATRES dataset \cite{ning2018multi}. Additionally, to evaluate the full potential of the proposed approach, we construct another dataset with more frequent event-timex interactions using distant supervision. On this dataset, our proposed approach substantially outperforms the ELMo-equipped baseline model.

\section{Methodology}
\label{sec:model}
\begin{figure}
\centering
    \includegraphics[trim=15mm 20mm 0mm 20mm,scale=0.34]{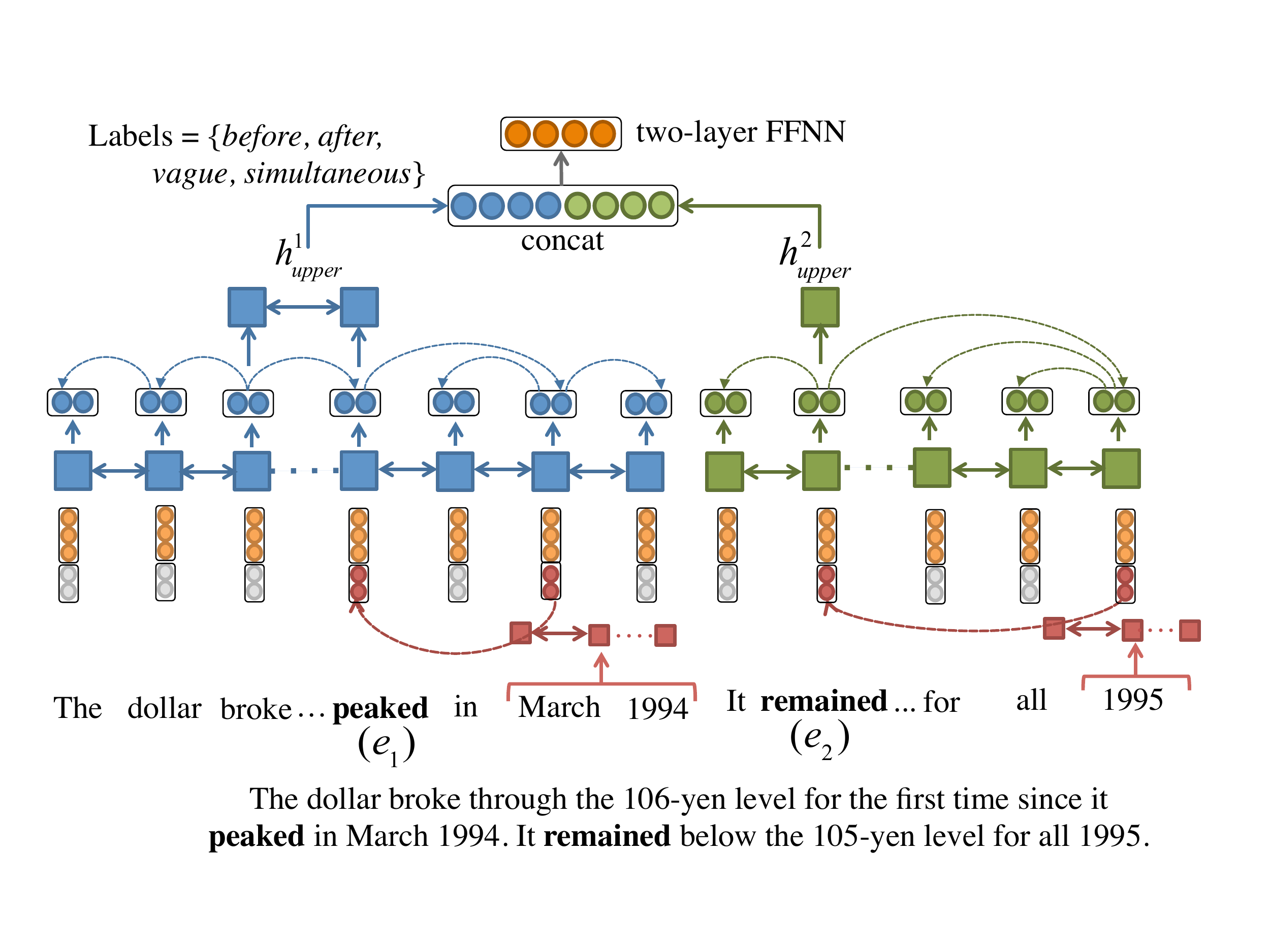}
    \caption{Temporal relation extraction model. Here, \textit{peaked} and \textit{remained} are the two events under consideration. The sentences are passed through the lower LSTM, then the outputs corresponding to the events' dependency paths are fed to the upper LSTMs, which produce input to feedforward and classification layers. Time expressions are embedded with a character-level model and broadcasted to events that they modify.}
    \label{fig:temprel-model}
\end{figure}

We improve upon the model architecture proposed by \citet{cheng2017classifying} for temporal relation extraction, which involves classifying the temporal relation between a given pair of events $e_1$ and $e_2$. Our proposed architecture is outlined in Figure \ref{fig:temprel-model}. The input to the system consists of two sentences, $s_1 = \{x^1_1, x^1_2,...x^1_n\}$ and $s_2 = \{x^2_1, x^2_2, ... x^2_m\}$ containing $e_1$ and $e_2$ respectively. Note that $s_1$ and $s_2$ may correspond to the same sentence.

\paragraph{Input Encoding} For each token $x_k$ in each sentence, we obtain a distributed representation $\tilde{x}_k = [v_w; v_p; v_t]$. Here, $v_w$ is the word embedding obtained from GloVe or contextualized word embeddings from ELMo, $v_p$ is a randomly initialized and trainable embedding of the part-of-speech tag, and $v_t$ corresponds to the timex embedding derived for time expressions (explained in Section \ref{sec:time-embed}).

\paragraph{Contextual Encoding} A biLSTM  is used to obtain contextualized embeddings $h_k$ for each token $x_k$ in the two sentences, as shown in Figure~\ref{fig:temprel-model}. The parameters are shared between these lower biLSTMs for the two sentences. Prior work \cite{cheng2017classifying} does not include these lower biLSTMs and only leverages the dependency encoding, explained next.

\paragraph{Dependency Encoding}
We use the Stanford Dependency Parser \cite{manning2014stanford} to extract the dependency paths for both events to their lowest common ancestor. For inter-sentence event pairs, paths are extracted to the root of each sentence. Each vector along the dependency path is fed into an upper biLSTM to produce output $h_{upper}$. Formally, for sentence $s_1$, 
\begin{equation*}
    h_{upper}^1 = \textrm{biLSTM}([h_k \textrm{ for } k \in \textrm{dep-path}(e_1)])
\end{equation*}
Parameters are shared between the upper biLSTMs for the two sentences. 

\paragraph{Output} We concatenate the outputs of the upper biLSTMs' embeddings for the two events to obtain $z = [h_{upper}^1; h_{upper}^2]$. We apply multiple feedforward layers with ReLU non-linearity, followed by a softmax layer to obtain output probabilities for the four labels \textit{before, after, vague} and \textit{simultaneous},\footnote{These are the labels used in the MATRES dataset \cite{ning2018improving}, but our classifier could in principle generalize to other label schemes as well.} denoting the temporal relation between the event pair $(e_1, e_2)$. The network is trained using the cross entropy loss.

\subsection{Time Embeddings}
Next, we outline our approach for constructing the timex embeddings $v_t$, which are concatenated to word and POS embeddings to generate the input encoding (as discussed in the previous section).

\paragraph{Training Data} To obtain time embeddings, we first constructed a grammar of time expressions in the dataset. We identified two main classes of timexes: explicit datetimes expressed in recognizable timex format (e.g. \emph{Sept.~12}, \emph{1993}, \emph{August 2013}, \emph{1998}, \emph{10-12-2014}, \emph{9th January}, etc.) and natural language time indicators (e.g.~\emph{two months ago}, \emph{5 weeks ago}, \emph{next year}, etc.). We designed generic templates that covered both these categories of timexes, e.g.~\lbrack
\textit{mm dd, yy}\rbrack.\footnote{See the appendix for more examples.} By randomly sampling values for the slots, we can generate valid time expressions based on this template. 
We used pairs of such randomly generated timexes to construct training data for our timex model. Since we generate time expression pairs from a pre-defined grammar and set of templates, it is straightforward to obtain the temporal order between the pairs of timexes. % In our formulation, we restrict to timex pairs with either \textit{before}, \textit{after} or \textit{simultaneous} as the temporal relation. \\

\begin{figure}
\centering
    \includegraphics[trim=20mm 15mm 10mm 93mm,scale=0.40]{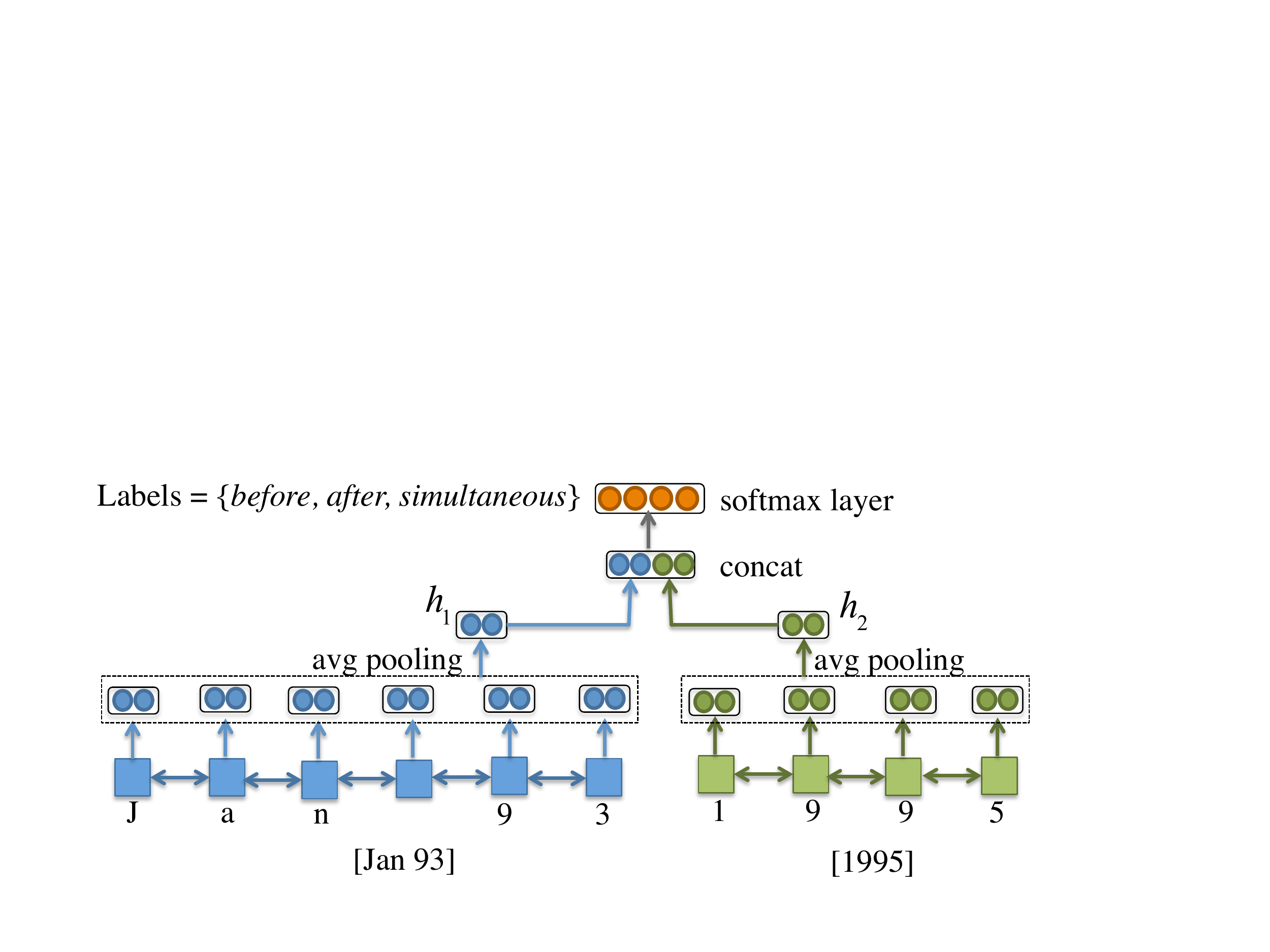}
    \caption{Timex model. The output of character biLSTMs is used to as input to classification. These vectors serve as time embeddings in the downstream tasks.}
    \label{fig:timex}
\end{figure}

\paragraph{Model Architecture} The model architecture for the timex model is outlined in Figure \ref{fig:timex}. The input to the system are two time expressions, $t_1$ and $t_2$. We use character biLSTMs to obtain distributed representations of both time expressions. We obtain time embeddings $h_1$ and $h_2$ for timex $t_1$ and $t_2$ by averaging the outputs of biLSTM layer. The two time embeddings are concatenated and fed through multiple feed forward layers with non-linearity. This is followed by a softmax layer that produces the output probabilities for the three label classes (\textit{before, after} and \textit{simultaneous}), denoting the temporal relation between the two time expressions. We train this network with the cross entropy loss. 

\paragraph{Inclusion in Temporal Models} For a given time expression, the average of the outputs of the bi-LSTM model ($h_1$) is used as the time embedding as shown in Figure~\ref{fig:temprel-model}.
%To incorporate this into temporal models, we concatenate the time embedding to the word embeddings and POS embeddings for time expressions. 
For other non-timex tokens, a zero vector is concatenated instead. Further, we also project the time embedding for a timex to the corresponding event it modifies according to a set of grammatical rules on the dependency parse, shown with red arrows in Figure~\ref{fig:temprel-model}.
\label{sec:time-embed}
\section{Experiments}

\subsection{Timex Pair Ordering}
First, we intrinsically evaluate the performance of the character-level timex model, outlined in Section \ref{sec:time-embed}. We generated $50000$ random pairs of time expressions for training and $5000$ randomly generated pairs for test. We seek to answer two questions: first, can our proposed timex model successfully capture temporal information necessary to order these timex pairs, and second, how effective are pre-trained embeddings for this task?

\begin{table}[]
\centering
\small
\begin{tabular}{c|c|c}
\toprule
%Model & \multicolumn{2}{|c}{Acc}  \\ \midrule
Model & w/ linear & w/ biLSTM \\ \midrule
GloVe embedding & $81.3$ & $88.7$\\
ELMo embedding & $88.3$ & $97.6$\\  
Char embedding (Ours) & $-$ & $97.3$\\  
\bottomrule
\end{tabular}
\caption{Performance on the synthetic timex dataset, classifying a pair of timexes as \emph{before}, \emph{after}, or \emph{simultaneous}. Including a biLSTM layer (as depicted in Figure~\ref{fig:timex}) leads to higher performance than just pooling and a linear layer. Character-level modeling (from ELMo or our learned embeddings) is important for high performance.}
\label{table:synthetic-timex}
\end{table} 

Table~\ref{table:synthetic-timex} shows a comparison between several models in our synthetic timex setting. Our proposed timex model achieves an accuracy of $97.3\%$. This high accuracy indicates that the model has effectively learned from the training data; its timex embeddings contain temporal ordering information which can be used for downstream tasks.

We also evaluate whether pre-trained embeddings such as ELMo or GloVe contain the necessary temporal information necessary for classifying the temporal order between timex pairs. We first test these with a minimal model. We construct a distributed representation of each time expression (obtained by average pooling the token level GloVe or ELMo embeddings), perform element-wise subtraction between the two embeddings, and feed the result through a linear classification layer that produces the output probabilities for the temporal label classes. The left column of Table~\ref{table:synthetic-timex} shows that while both GloVe and ELMo contain some temporal information, our proposed model's additional parameters and richer embedding scheme lead to higher performance.

We further experiments to investigate if ELMo or GloVe can additionally be used in our timex model to obtain even more powerful embeddings. We replace our model's character-level vectors and character-level biLSTM with token-level pretrained vectors (either contextualized vectors from ELMo or non-contextual vectors from GloVe) and a token-level biLSTM. As before, the outputs of this biLSTM for the two timexes are concatenated and further fed to feedforward and softmax layers for temporal label prediction. Using ELMo embeddings in this manner does not lead to a substantial improvement over previous results, with an accuracy of $97.6\%$ for the temporal relation classification objective on the same test set. However, the performance using GloVe embeddings drops to $88.7\%$. This drop in performance can partially be attributed to the word-level nature of GloVe vectors, which do not necessarily cover every year that might be seen in the dataset. We used the GloVe vectors with $840$ billion tokens (largest available) to circumvent this issue and minimize the number of out of vocabulary instances, but still see low performance.

\subsection{Event Temporal Ordering}

Next, we investigate the effectiveness of our timex embeddings in the context of our full event temporal ordering model. We evaluate on two event temporal ordering datasets, one real and one artificially constructed.

\subsubsection{Evaluation on MATRES}

We evaluate on the MATRES dataset proposed in \citet{ning2018multi}. This dataset is designed to be less ambiguous than TimeBank-Dense \cite{cassidy2014annotation}. MATRES contains temporal annotations for documents from the TimeBank \cite{pustejovsky2003timebank}, AQUAINT \cite{graff2002aquaint} and Platinum datasets \cite{uzzaman2013semeval}. We follow standard practice and use TimeBank and AQUAINT ($256$ articles) for training and Platinum ($20$ articles) for testing.

Table \ref{table:matres-timex} outlines the performance of the proposed approach on MATRES. We implemented the model proposed by \citet{cheng2017classifying} and compare against it. We evaluate the models using both GloVe and ELMo embeddings. Our results show substantial improvement over this baseline model.  Moreover, including time embeddings as additional input to the improved models leads to a small improvement in the overall accuracy. However, we did not find the results to be statistically significant according to a bootstrap resampling test (GloVe $p$-value $= 0.349$, ELMo $p$-value $= 0.267$).\footnote{Augmenting word embeddings with time embeddings increases the number of network parameters; however, additional experiments revealed that increasing the size of the GloVe embeddings in the basic temporal model did not lead to an improvement in performance. Therefore, it does not seem that extra parameters in the model contribute to the observed improvements.}

Note that only a fraction of examples in the MATRES dataset contain distinct time expressions that can be compared to resolve temporal ordering. To further evaluate our approach, we investigated whether an equivalent performance improvement could be achieved through post-processing rules involving time expressions. We identified event pairs in the data for which both events had an accompanying time expression modifying the event according to the dependency parse. We can then infer the temporal relation between the event pair using rules on top of these timexes. However, we observed that such a post-processing scheme had very low coverage in the dataset and could not repair \emph{any} errors in the development set. We therefore turn our attention to a setting with a richer set of timexes for further evaluation.\footnote{In prior work \cite{cheng2017classifying,  meng2018context}, machine learning classifiers are used to infer a wider range of event-timex links, which can potentially increase the informativeness of timexes. However, many of the links they target require complex inferences to determine, and as a result those works report relatively low performance for such classifiers. Hence, we do not compare to these methods in our experiments.}

\begin{table}[]
\centering
\small
\begin{tabular}{c|c|c}
\toprule
%Model & \multicolumn{2}{|c}{Acc}  \\ \midrule
Model & GloVe & ELMo \\ \midrule
\newcite{cheng2017classifying} & $59.53$ & $65.50$\\
Ours w/o timex embed & $62.83$ & $68.45$\\  
Ours w/ timex embed & $63.22$ & $68.61$\\  
\bottomrule
\end{tabular}
\caption{Performance of our event temporal ordering model on the MATRES dataset. We report the mean accuracy over 3 runs of each model. Our model improves substantially over \newcite{cheng2017classifying}. Including timexes leads to small accuracy gains, partially due to the fact that timexes often do not occur with the dataset's hard examples.}
\label{table:matres-timex}
\end{table}

\subsubsection{Evaluation on Distant Data}
In MATRES, only a fraction of the examples contain time expressions and are consequently affected by inclusion of time embeddings. Therefore, to test the full potential of the proposed approach, we additionally collect a test dataset of examples with explicit timexes that expose their temporal relation; we view the timexes as distant supervision for the event pairs. To identify such examples, we use two high precision classifiers proposed in \citet{chambers2014dense}: (a) an event-timex classifier that identifies the temporal relation between adjacent verb and time expressions (precision = $0.92$), (b) a timex-timex classifier that identifies the temporal relation between two time expressions (precision = $0.88$). These classifiers can allow us to directly infer the time relation between an event pair where each event is linked to a timex. An example event pair from the distant data thus collected is: \textit{``Riyadh \textbf{suspended} aid to the Palestinians in 1990 when it accused Arafat of siding with Iraq after the 1990 invasion of Kuwait, but it \textbf{restored} aid in 1994."}\footnote{See the appendix for more samples from the distant data.} Note that the classifiers used have very low recall in general, but by running the system on Gigaword \cite{GraffEtAl2007}, we can extract a large dataset in spite of this. 

\begin{table}[]
\centering
\small
\begin{tabular}{c|ccc}
\toprule
 & 2000 & 3000 & 4000 \\ \midrule
\multicolumn{4}{c}{GloVe} \\ \midrule
Ours w/o Timex Embed & $74.0$ & $76.8$ & $78.2$ \\
Ours w/ Masked Timex & $73.9$ & $75.5$ & $77.1$ \\ 
Ours w/ Timex Embed & $81.6$ & $83.2$ & $83.1$ \\ \midrule
\multicolumn{4}{c}{ELMo} \\ \midrule
Ours w/o Timex Embed & $80.1$ & $83.8$ & $84.3$ \\ 
Ours w/ Masked Timex & $79.8$ & $80.1$ & $80.7$ \\ 
Ours w/ Timex Embed & $82.3$ & $84.5$ & $84.8$ \\ \bottomrule
\end{tabular}
\caption{Performance of our models on the distantly-labeled event ordering data. We report overall accuracy values. In both the GloVe and ELMo settings, our timex embeddings lead to higher performance. The ELMo model gets substantially worse when timexes are masked, indicating that it is organically exploiting these better than GloVe is.}
\label{table:distant-data}
\end{table}

Since this distant data is created using rule-based classifiers, given a large amount of training data, the baseline model can achieve high performance as it learns to infer these rules. However, our aim is to improve the performance of the event ordering model on moderately sized datasets, where the knowledge induction from timex embeddings play a larger role. Therefore, we report results on training sets of size $2000$, $3000$, and $4000$ samples. The test set is kept constant with $1000$ samples.

Table \ref{table:distant-data} outlines the performance of the temporal models on this dataset. We evaluate our models across three settings: (a) our event ordering model without including timex embeddings, (b) our event ordering model with masking of time tokens (replacing it with UNK tokens) and (c) our full model including timex embeddings. 
%By masking out time tokens in setting b, the model learns to rely on signals other that those from timexes. Therefore, by comparing these results with those of setting a, we were able to determine whether the model attends to time expressions efficiently. 
We evaluate the models using both GloVe and ELMo embeddings as input. In both settings, incorporating our timexes leads to higher performance. For GloVe, the performance of the basic temporal model is similar to that when the time expression is masked out. This demonstrates that the temporal model does not use the knowledge from time expressions when making temporal relation predictions. However, in the ELMo setting, we observed a larger drop in performance by masking out the time expressions compared to GloVe embeddings. This demonstrates that the ELMo embeddings are not agnostic to time-expressions in the sentence, although they still show improvement by inclusion of timex embeddings trained specifically with the temporal classification objective on small datasets.  

\section{Conclusion}
In this paper, we propose a framework to learn temporally-aware timex embeddings from synthetic data. Through experiments on two datasets, we show that incorporating these embeddings in deep temporal models leads to an improvement in the overall temporal classification performance. 

\section{Acknowledgments}
This work was partially supported by NSF Grant IIS-1814522, NSF Grant SHF-1762299, a Bloomberg Data Science Grant, and an equipment grant from NVIDIA. The authors acknowledge the Texas Advanced Computing Center (TACC) at The University of Texas at Austin for providing HPC resources used to conduct this research. Also, thanks to the anonymous reviewers for their helpful comments.

\bibliography{acl2019}

\begin{thebibliography}{23}
\expandafter\ifx\csname natexlab\endcsname\relax\def\natexlab#1{#1}\fi

\bibitem[{Bramsen et~al.(2006)Bramsen, Deshpande, Lee, and Barzilay}]{Bramsen}
Philip Bramsen, Pawan Deshpande, Yoong~Keok Lee, and Regina Barzilay. 2006.
\newblock \href {http://dl.acm.org/citation.cfm?id=1610075.1610105} {{Inducing
  Temporal Graphs}}.
\newblock In \emph{Proceedings of the 2006 Conference on Empirical Methods in
  Natural Language Processing}, EMNLP '06, pages 189--198, Stroudsburg, PA,
  USA. Association for Computational Linguistics.

\bibitem[{Cassidy et~al.(2014)Cassidy, McDowell, Chambers, and
  Bethard}]{cassidy2014annotation}
Taylor Cassidy, Bill McDowell, Nathanael Chambers, and Steven Bethard. 2014.
\newblock {An Annotation Framework for Dense Event Ordering}.
\newblock In \emph{Proceedings of the 52nd Annual Meeting of the Association
  for Computational Linguistics (Volume 2: Short Papers)}, volume~2, pages
  501--506.

\bibitem[{Chambers et~al.(2014)Chambers, Cassidy, McDowell, and
  Bethard}]{chambers2014dense}
Nathanael Chambers, Taylor Cassidy, Bill McDowell, and Steven Bethard. 2014.
\newblock {Dense event ordering with a multi-pass architecture}.
\newblock \emph{Transactions of the Association for Computational Linguistics},
  2:273--284.

\bibitem[{Chambers and Jurafsky(2008)}]{Chambers:2008}
Nathanael Chambers and Dan Jurafsky. 2008.
\newblock \href {http://dl.acm.org/citation.cfm?id=1613715.1613803} {{Jointly
  Combining Implicit Constraints Improves Temporal Ordering}}.
\newblock In \emph{Proceedings of the Conference on Empirical Methods in
  Natural Language Processing}, EMNLP '08, pages 698--706, Stroudsburg, PA,
  USA. Association for Computational Linguistics.

\bibitem[{Cheng and Miyao(2017)}]{cheng2017classifying}
Fei Cheng and Yusuke Miyao. 2017.
\newblock {Classifying Temporal Relations by Bidirectional {LSTM} over
  Dependency Paths}.
\newblock In \emph{Proceedings of the 55th Annual Meeting of the Association
  for Computational Linguistics (Volume 2: Short Papers)}, volume~2, pages
  1--6.

\bibitem[{Do et~al.(2012)Do, Lu, and Roth}]{do2012joint}
Quang~Xuan Do, Wei Lu, and Dan Roth. 2012.
\newblock {Joint inference for event timeline construction}.
\newblock In \emph{Proceedings of the 2012 Joint Conference on Empirical
  Methods in Natural Language Processing and Computational Natural Language
  Learning}, pages 677--687. Association for Computational Linguistics.

\bibitem[{Graff(2002)}]{graff2002aquaint}
David Graff. 2002.
\newblock {The AQUAINT corpus of English news text}.
\newblock \emph{Linguistic Data Consortium, Philadelphia}.

\bibitem[{Graff et~al.(2007)Graff, Kong, Chen, and Maeda}]{GraffEtAl2007}
David Graff, Junbo Kong, Ke~Chen, and Kazuaki Maeda. 2007.
\newblock {English Gigaword Third Edition}.
\newblock Linguistic Data Consortium, Catalog Number LDC2007T07.

\bibitem[{Llorens et~al.(2015)Llorens, Chambers, UzZaman, Mostafazadeh, Allen,
  and Pustejovsky}]{llorens2015semeval}
Hector Llorens, Nathanael Chambers, Naushad UzZaman, Nasrin Mostafazadeh, James
  Allen, and James Pustejovsky. 2015.
\newblock {SemEval-2015 Task 5: QA TEMPEVAL-Evaluating Temporal Information
  Understanding with Question Answering}.
\newblock In \emph{Proceedings of the 9th International Workshop on Semantic
  Evaluation (SemEval 2015)}, pages 792--800.

\bibitem[{Manning et~al.(2014)Manning, Surdeanu, Bauer, Finkel, Bethard, and
  McClosky}]{manning2014stanford}
Christopher Manning, Mihai Surdeanu, John Bauer, Jenny Finkel, Steven Bethard,
  and David McClosky. 2014.
\newblock {The Stanford CoreNLP natural language processing toolkit}.
\newblock In \emph{Proceedings of 52nd annual meeting of the association for
  computational linguistics: system demonstrations}, pages 55--60.

\bibitem[{Meng and Rumshisky(2018)}]{meng2018context}
Yuanliang Meng and Anna Rumshisky. 2018.
\newblock {Context-Aware Neural Model for Temporal Information Extraction}.
\newblock In \emph{Proceedings of the 56th Annual Meeting of the Association
  for Computational Linguistics (Volume 1: Long Papers)}, volume~1, pages
  527--536.

\bibitem[{Meng et~al.(2017)Meng, Rumshisky, and Romanov}]{meng2017temporal}
Yuanliang Meng, Anna Rumshisky, and Alexey Romanov. 2017.
\newblock {Temporal Information Extraction for Question Answering Using
  Syntactic Dependencies in an LSTM-based Architecture}.
\newblock In \emph{Proceedings of the 2017 Conference on Empirical Methods in
  Natural Language Processing}, pages 887--896.

\bibitem[{Mikolov et~al.(2013)Mikolov, Sutskever, Chen, Corrado, and
  Dean}]{mikolov2013distributed}
Tomas Mikolov, Ilya Sutskever, Kai Chen, Greg~S Corrado, and Jeff Dean. 2013.
\newblock {Distributed representations of words and phrases and their
  compositionality}.
\newblock In \emph{Advances in neural information processing systems}, pages
  3111--3119.

\bibitem[{Mostafazadeh et~al.(2016)Mostafazadeh, Grealish, Chambers, Allen, and
  Vanderwende}]{mostafazadeh2016caters}
Nasrin Mostafazadeh, Alyson Grealish, Nathanael Chambers, James Allen, and Lucy
  Vanderwende. 2016.
\newblock {CaTeRS: Causal and temporal relation scheme for semantic annotation
  of event structures}.
\newblock In \emph{Proceedings of the Fourth Workshop on Events}, pages 51--61.

\bibitem[{Ning et~al.(2017)Ning, Feng, and Roth}]{ning2017structured}
Qiang Ning, Zhili Feng, and Dan Roth. 2017.
\newblock {A structured learning approach to temporal relation extraction}.
\newblock In \emph{Proceedings of the 2017 Conference on Empirical Methods in
  Natural Language Processing}, pages 1027--1037.

\bibitem[{Ning et~al.(2018{\natexlab{a}})Ning, Feng, Wu, and
  Roth}]{ning2018joint}
Qiang Ning, Zhili Feng, Hao Wu, and Dan Roth. 2018{\natexlab{a}}.
\newblock {Joint reasoning for temporal and causal relations}.
\newblock In \emph{Proceedings of the 56th Annual Meeting of the Association
  for Computational Linguistics (Volume 1: Long Papers)}, volume~1, pages
  2278--2288.

\bibitem[{Ning et~al.(2018{\natexlab{b}})Ning, Wu, Peng, and
  Roth}]{ning2018improving}
Qiang Ning, Hao Wu, Haoruo Peng, and Dan Roth. 2018{\natexlab{b}}.
\newblock {Improving Temporal Relation Extraction with a Globally Acquired
  Statistical Resource}.
\newblock In \emph{Proceedings of the 2018 Conference of the North American
  Chapter of the Association for Computational Linguistics: Human Language
  Technologies, Volume 1 (Long Papers)}, pages 841--851.

\bibitem[{Ning et~al.(2018{\natexlab{c}})Ning, Wu, and Roth}]{ning2018multi}
Qiang Ning, Hao Wu, and Dan Roth. 2018{\natexlab{c}}.
\newblock {A Multi-Axis Annotation Scheme for Event Temporal Relations}.
\newblock In \emph{Proceedings of the 56th Annual Meeting of the Association
  for Computational Linguistics (Volume 1: Long Papers)}, pages 1318--1328.

\bibitem[{Pennington et~al.(2014)Pennington, Socher, and
  Manning}]{pennington2014glove}
Jeffrey Pennington, Richard Socher, and Christopher Manning. 2014.
\newblock {Glove: Global vectors for word representation}.
\newblock In \emph{Proceedings of the 2014 conference on empirical methods in
  natural language processing (EMNLP)}, pages 1532--1543.

\bibitem[{Peters et~al.(2018)Peters, Neumann, Iyyer, Gardner, Clark, Lee, and
  Zettlemoyer}]{Peters:2018}
Matthew Peters, Mark Neumann, Mohit Iyyer, Matt Gardner, Christopher Clark,
  Kenton Lee, and Luke Zettlemoyer. 2018.
\newblock {Deep Contextualized Word Representations}.
\newblock In \emph{Proceedings of the 2018 Conference of the North American
  Chapter of the Association for Computational Linguistics: Human Language
  Technologies, Volume 1 (Long Papers)}, pages 2227--2237.

\bibitem[{Pustejovsky et~al.(2003)Pustejovsky, Hanks, Sauri, See, Gaizauskas,
  Setzer, Radev, Sundheim, Day, Ferro et~al.}]{pustejovsky2003timebank}
James Pustejovsky, Patrick Hanks, Roser Sauri, Andrew See, Robert Gaizauskas,
  Andrea Setzer, Dragomir Radev, Beth Sundheim, David Day, Lisa Ferro, et~al.
  2003.
\newblock {The timebank corpus}.
\newblock In \emph{Corpus linguistics}, volume 2003, page~40. Lancaster, UK.

\bibitem[{Tourille et~al.(2017)Tourille, Ferret, Neveol, and
  Tannier}]{tourille2017neural}
Julien Tourille, Olivier Ferret, Aurelie Neveol, and Xavier Tannier. 2017.
\newblock {Neural Architecture for Temporal Relation Extraction: A Bi-LSTM
  Approach for Detecting Narrative Containers}.
\newblock In \emph{Proceedings of the 55th Annual Meeting of the Association
  for Computational Linguistics (Volume 2: Short Papers)}, volume~2, pages
  224--230.

\bibitem[{UzZaman et~al.(2013)UzZaman, Llorens, Derczynski, Allen, Verhagen,
  and Pustejovsky}]{uzzaman2013semeval}
Naushad UzZaman, Hector Llorens, Leon Derczynski, James Allen, Marc Verhagen,
  and James Pustejovsky. 2013.
\newblock {Semeval-2013 task 1: Tempeval-3: Evaluating time expressions,
  events, and temporal relations}.
\newblock In \emph{Second Joint Conference on Lexical and Computational
  Semantics (* SEM), Volume 2: Proceedings of the Seventh International
  Workshop on Semantic Evaluation (SemEval 2013)}, volume~2, pages 1--9.

\end{thebibliography}
\bibliographystyle{acl_natbib}

\appendix

\section{Appendix}

\begin{table*}
\begin{tabular}{p{14cm}|c}
\hline
\multicolumn{1}{c}{\textbf{Event Pair}} & \multicolumn{1}{c}{\textbf{Label}} \\ \hline
Former Singapore premier Lee Kuan Yew, who \textbf{came} to power in 1959, \textbf{stepped} down in 1990 in favour of the incumbent, prime minister Goh Chok Tong, but remains influential as a senior minister in Goh's cabinet. &  Before \\ \hline
Relations between Sudan and Saudi Arabia \textbf{grew} tense in 1990 when Riyadh accused Khartoum of supporting Iraq after its invasion of Kuwait and \textbf{worsened} in 1992 when Sudan granted asylum to Saudi militant Osama Bin Laden. &  Before\\ \hline
The Israeli-Syrian peace talks \textbf{launched} in 1991 are mainly focusing on Damascus' insistence that Israel withdraw its troops from the Golan Heights in exchange for peace. That territory has been \textbf{occupied} by Israeli troops since 1967. & After \\ \hline
Resolutions were \textbf{passed} by the UN Security Council after the first Indo-Pakistan war over Kashmir in 1948. The dispute \textbf{led} to a second war between the neighbours in 1965. &  Before\\ \hline
Turkish mainland forces \textbf{invaded} Northern Cyprus in 1974 after a coup in Nicosia backed by the military junta then ruling Greece. A Turkish-Cypriot state was \textbf{declared} in 1983, and Ankara now has about 35,000 troops and 400 tanks stationed there. &  Before\\ \hline
More people \textbf{watched} Formula One on television in 1995 than \textbf{watched} the world cup in 1994. & After\\ \hline
He was \textbf{freed} six months early in September 1993 but \textbf{re-arrested} in April 1994 after meeting with John Shattuck, the US assistant secretary of state for human rights. &  Before\\ \hline
\end{tabular}
\caption{Examples from the distantly-labeled event ordering data. Events are shown in bold and may be co-located in a single sentence or span two sentences. Event-timex relations are recognized with high-precision classifiers from \citet{chambers2014dense}.}
\label{table:distant-data-ex}
\end{table*}

\subsection{Timex Templates}
We use generic templates for time expressions to generate training data for the timex model. Two kinds of templates were generated: (1) explicit datetimes, and (2) natural language time indicators. Examples of each of these kinds are outlined below:
\begin{enumerate}
    \item Explicit datetime templates: \textit{ \lbrack yyyy\rbrack, \lbrack 'yy\rbrack, \lbrack mm dd yy\rbrack, \lbrack mm yy\rbrack, \lbrack mmm yyyy\rbrack, \lbrack mmm dd yyyy\rbrack}, etc.
    \item Natural language indicators: \textit{\lbrack xx units later\rbrack, \lbrack xx units before\rbrack, \lbrack now\rbrack, \lbrack past xx units\rbrack}, etc., where \textit{xx} is filled by a numerical value and \textit{units} refers to a time unit such as months, days, or years. 
\end{enumerate}

Timex pairs generated through these templates can be converted to a standardized time scale and hence easily compared. It is therefore straight forward to infer the gold label for each pair of generated timexes. For MATRES, $75\%$ of the pairs in the training set for the timex model are sampled from explicit datetime templates, and the rest are sampled from natural language templates. This relative ratio was heuristically determined. $100\%$ of the pairs were drawn from explicit datetime templates for the distant data.

\subsection{Examples from Distant Data}
Table \ref{table:distant-data-ex} provides some examples of event pairs, and their corresponding label from the distant data. This dataset is automatically created using two high precision rule-based classifiers.

\end{document}